\documentclass[lettersize,journal]{IEEEtran}
\usepackage{amsmath,amsfonts}
\usepackage{algorithmic}
\usepackage{array}
\usepackage[caption=false,font=normalsize,labelfont=sf,textfont=sf]{subfig}
\usepackage{textcomp}
\usepackage{stfloats}
\usepackage{url}
\usepackage{verbatim}
\usepackage{graphicx}
\def\BibTeX{{\rm B\kern-.05em{\sc i\kern-.025em b}\kern-.08em
    T\kern-.1667em\lower.7ex\hbox{E}\kern-.125emX}}
\usepackage{balance}

\usepackage[colorlinks,urlcolor=blue,linkcolor=blue,citecolor=blue]{hyperref}

\usepackage{color,array}
\usepackage{amsmath}
\usepackage{graphicx}
\usepackage{booktabs}
\usepackage{multirow}
\usepackage{subcaption}
\usepackage{orcidlink}
\usepackage{algorithm}
\usepackage{algorithmic}
\usepackage{balance}
\usepackage{amssymb}

\usepackage[normalem]{ulem}
\useunder{\uline}{\ul}{}

\begin{document}

\title{Attribution for Enhanced Explanation with Transferable Adversarial eXploration}

\author{Zhiyu Zhu\orcidlink{0009-0009-0231-4410}\textsuperscript{*}, Jiayu Zhang\orcidlink{0009-0008-6636-8656}\textsuperscript{*}, Zhibo Jin\orcidlink{0009-0003-0218-1941}, Huaming Chen\orcidlink{0000-0001-5678-472X}, Jianlong Zhou\orcidlink{0000-0001-6034-644X} and Fang Chen\orcidlink{0000-0003-4971-8729}
% \thanks{This paragraph of the first footnote will contain the date on which you submitted your paper for review. It will also contain support information, including sponsor and financial support acknowledgment. For example, ``This work was supported in part by the U.S. Department of Commerce under Grant BS123456.'' }
\thanks{\textsuperscript{*}Zhiyu Zhu and Jiayu Zhang contributed equally to this work.}
\thanks{Zhiyu Zhu, Zhibo Jin, Jianlong Zhou and Fang Chen are with the School of Computer Science, Faculty of Engineering and IT, University of Technology Sydney, Australia. (e-mail: zhiyu.zhu@student.uts.edu.au).}%Xinyi Wang is with Faculty of Computer Science and Information Technology, University of Malaya, Malaysia. (e-mail: 22103906@siswa.um.edu.my). Jiayu Zhang is with SuZhou Yierqi, China. (e-mail: zjy@szyierqi.com). Kim-Kwang Raymond Choo is with The University of Texas at San Antonio, San Antonio, TX 78249, USA. (e-mail:raymond.choo@fulbrightmail.org). * is the corresponding author.}
% \thanks{S. B. Author, Jr., was with Rice University, Houston, TX 77005 USA. He is now with the Department of Physics, Colorado State University, Fort Collins, CO 80523 USA (e-mail: author@lamar.colostate.edu).}
% \thanks{T. C. Author is with the Electrical Engineering Department, University of Colorado, Boulder, CO 80309 USA, on leave from the National Research Institute for Metals, Tsukuba, Japan (e-mail: author@nrim.go.jp).}
% \thanks{This paragraph will include the Associate Editor who handled your paper.}
\thanks{This work has been submitted to the IEEE for possible publication. Copyright may be transferred without notice, after which this version may no longer be accessible.}
}

\markboth{}
{}

\maketitle

\begin{abstract}
The interpretability of deep neural networks is crucial for understanding model decisions in various applications, including computer vision. AttEXplore++, an advanced framework built upon AttEXplore, enhances attribution by incorporating transferable adversarial attack methods such as MIG and GRA, significantly improving the accuracy and robustness of model explanations. We conduct extensive experiments on five models, including CNNs (Inception-v3, ResNet-50, VGG16) and vision transformers (MaxViT-T, ViT-B/16), using the ImageNet dataset. Our method achieves an average performance improvement of 7.57\% over AttEXplore and 32.62\% compared to other state-of-the-art interpretability algorithms. Using insertion and deletion scores as evaluation metrics, we show that adversarial transferability plays a vital role in enhancing attribution results. Furthermore, we explore the impact of randomness, perturbation rate, noise amplitude, and diversity probability on attribution performance, demonstrating that AttEXplore++ provides more stable and reliable explanations across various models.
 We release our code at: \hyperlink{https://anonymous.4open.science/r/ATTEXPLOREP-8435/}{https://anonymous.4open.science/r/ATTEXPLOREP-8435/}

%Transferable adversarial attacks, given their Feature-level attacks disrupt the fundamental components of the model by exploiting estimates of the importance of intermediate layer neurons, thereby generating transferable adversarial samples. However, Constrained by the properties of attribution integration paths and the chosen baseline points, attribution-based methods struggle to provide promising feature attributions across multiple target models. Building on our previous work DANAA, in this paper, we reveal the similarity in attributions across different models. Secondly, we delve into the impact of combining different attribution properties on attribution for sample transferability. By utilizing various adversarial attack methods to generate different adversarially trained baseline points and performing attribution along linear or nonlinear integration paths, we propose a novel neuron attribution-based attack, DANAA++. Finally, we find that the aggressiveness of baseline points and the nonlinearity of integration paths both significantly affect sample transferability, with the aggressiveness of baseline points playing a more critical role. Our approach provides insights for building superior attribution-based feature-level attacks in the future. Our code is available at:
\end{abstract}

% \begin{IEEEImpStatement}
% Current black-box adversarial attack methods remain challenging to manipulate the deep learning model's behaviour, in particular for the ones with defense mechanisms. In this work, a novel feature-level transferable attack approach is presented to generate more effective adversarial samples by exploring different properties of attribution methods. By highlighting the attribution outcome for neuron importance estimation, the proposed method provides an innovative perspective for adversarial attacks based on the intermediate layer neurons. It can be widely applied in various safety-critical attack scenarios, including autonomous driving, medical diagnosis, financial analysis and so on. 
% In addition, we anticipate our method will provide further insight into the robustness testing of machine learning models to effectively ensure user information security.

% \end{IEEEImpStatement}


\section{Introduction}
\IEEEPARstart{W}{ith} the widespread application of Deep Neural Networks (DNNs) in critical fields such as medical diagnostics, autonomous driving, and financial forecasting, the interpretability of their decision-making processes has become an essential research direction~\cite{lecun2015deep,esteva2017dermatologist,chen2016xgboost}. Although DNN models demonstrate excellent performance across various complex tasks, their black-box nature limits our understanding of their internal workings~\cite{doshi2017towards,lipton2018mythos,rudin2019stop}. This lack of transparency not only hinders users' trust in model decisions but also complicates the evaluation and correction of models in real-world applications~\cite{goodfellow2014explaining}, particularly in domains with high security and fairness requirements~\cite{mehrabi2021survey}.

The goal of interpretability methods is to enhance the transparency of DNNs by revealing how the models derive decisions from input features~\cite{somani2024propagating}. Among these, attribution methods have garnered significant attention for their ability to provide fine-grained explanations on a per-feature basis~\cite{hooker2019benchmark}. Attribution methods calculate the contribution of each input feature to the model's output, generating visual heatmaps that clearly show which features are most important for the model's decisions~\cite{montavon2018methods,tjoa2020survey}. These methods not only help understand the model’s decision process but also identify potential biases or vulnerabilities~\cite{thibeau2023interpretability}.

At the same time, adversarial attacks play an important role in interpretability research. By applying small perturbations to model inputs, adversarial attacks can approximate and explore the model's decision boundaries~\cite{narodytska2017simple,wang2022di}. Adversarial examples often reveal which features are critical in the model's decision-making process because adversarial attacks aim to identify and exploit weaknesses in the model~\cite{ross2018improving,subramanya2019fooling}. These vulnerabilities frequently correspond to the instability of the model in certain feature dimensions, making adversarial examples a valuable tool for identifying the most sensitive features in the model’s decisions~\cite{zhang2021adversarial}. This capability positions adversarial attacks as an effective means of exploring and enhancing DNN interpretability, providing a unique perspective into the internal mechanisms of models~\cite{han2023interpreting}.

Current research has proposed the AttEXplore~\cite{zhuattexplore} method, which focuses on enhancing the interpretability of attribution methods through transferability. AttEXplore leverages a mechanism of transferability across models and tasks, enabling it to not only identify important features in the source model but also maintain the interpretability and stability of these features in the target model. By combining attribution analysis with transfer learning, AttEXplore aims to reveal feature importance across different environments, expanding the applicability of interpretability methods. AttEXplore explores both the frequency and spatial domains of input features to generate transferable examples that help in understanding the features. These examples not only adjust model parameters but also maintain the validity and stability of features when transferred to new models or tasks. However, AttEXplore mainly relies on a limited number of adversarial attack methods and does not fully utilize more extensive and newer methods of adversarial transferability. Furthermore, AttEXplore does not deeply consider the impact of randomness in the attribution process, which could lead to issues in the stability and reliability of the results.

\begin{figure*}
    \centering
    \includegraphics[width=0.65\linewidth]{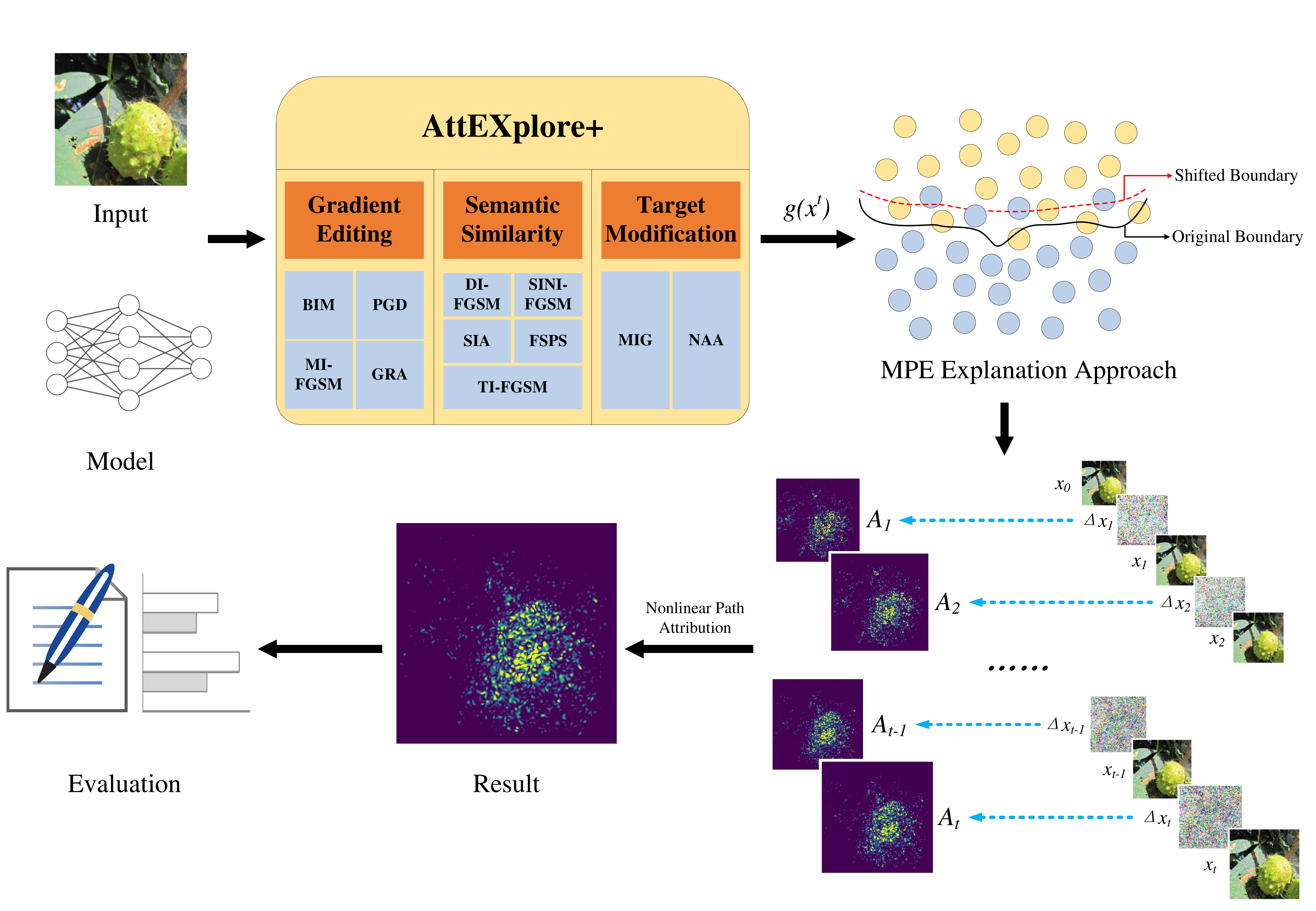}
    \caption{Flowchart of AttEXplore+ Framework}
    \label{fig:enter-label}
\end{figure*}
We further extend AttEXplore and propose AttEXplore+, which explores various types of transferable adversarial attack methods, addressing the limitations of AttEXplore in terms of attack diversity and stability. AttEXplore+ integrates 10 methods from three major categories of transferable adversarial attacks, significantly broadening and deepening the exploration of model decision boundaries. These methods include Gradient Editing, Semantic Similarity, Target Modification, and other attack types, generating a broader range of adversarial examples to provide more detailed information for attribution analysis. Through comparative experiments, we observe that different attack types vary in their effectiveness in explaining feature importance in models, thus demonstrating the value of diverse attacks in improving interpretability.

Moreover, AttEXplore+ expands not only the types of attacks but also specifically addresses the impact of randomness on attribution results. Through more in-depth ablation studies, we investigate the role of different levels of randomness in generating adversarial examples and performing attribution analysis. We explore how to leverage and control randomness to enhance the robustness and reliability of results. These experiments show that introducing an appropriate level of randomness can effectively improve the transferability and stability of attribution methods across different models and tasks.

The main contributions and research questions of this study include:
\begin{itemize}
    \item How do different adversarial transfer methods affect the performance of AttEXplore+? We explore the effectiveness of various adversarial methods in revealing model decisions and compare their performance in attribution results through analysis and experiments.
    \item What role do the parameters of large-scale adversarial attack methods play in the stability and reliability of adversarial attribution results? We examine the impact of parameters such as perturbation strength, optimization initialization, and randomness control on the attribution performance of AttEXplore+.
    \item We propose an optimized version of the AttEXplore+ framework, named AttEXplore++, which achieves the best attribution performance.
    \item We present a new attribution framework named AttEXplore+ that can be optimized using other adversarial attack methods capable of computing gradient information. The code will be open-sourced to facilitate usage and continued research by relevant researchers and developers.
\end{itemize}

\section{Related Works}

\subsection{Local Approximation Methods}
Local approximation methods aim to simplify the model's complexity around individual data points, providing local explanations for globally complex models. These methods are widely applicable as they do not rely on the internal structure of the target model. For example, LIME (Local Interpretable Model-Agnostic Explanations)~\cite{ribeiro2016should} uses local linear approximations to explain models, making it particularly suitable for handling nonlinear or non-differentiable models. However, the accuracy of LIME depends on the local linearity assumption and the selection of perturbations, which may not adequately capture the global behavior of the model. While submodular optimization can be used to select more representative samples to improve global explanations, LIME is more appropriate for local interpretations. Additionally, LIME requires training a local linear model for each explanation instance, which can be challenging when handling large datasets or complex models. In contrast, SHAP (Shapley Additive Explanations)~\cite{lundberg2017unified} addresses the lack of global consistency in LIME by allocating fair contributions to each feature based on Shapley values. However, SHAP suffers from computational inefficiency due to the need to calculate the marginal contribution of each feature, which becomes computationally expensive as the input dimension increases exponentially. Another method, DeepLIFT~\cite{shrikumar2017learning}, attributes the change in model output to input features by propagating neuron activation differences, effectively avoiding gradient saturation problems, but it does not satisfy the implementation invariance axiom~\cite{sundararajan2017axiomatic}.

\subsection{Gradient-Based Attribution Methods}
Gradient-based attribution methods extend the idea of local approximation methods by calculating the gradients of model outputs with respect to inputs, providing finer-grained explanations. Saliency Maps (SM) were initially introduced as a visualization tool for deep convolutional neural networks to highlight which input features (such as pixels) most influence the model's predictions~\cite{simonyan2013deep}. However, SM has limitations in handling gradient noise and gradient saturation issues, and it does not fully satisfy two important attribution axioms: Sensitivity and Implementation Invariance.

To address these issues, Integrated Gradients (IG) was introduced to satisfy the aforementioned two attribution axioms, greatly advancing the field of attribution methods. The Sensitivity axiom requires that if a change in an input feature leads to a change in the model output, that feature should receive a non-zero attribution. The Implementation Invariance axiom mandates that attribution results should depend solely on the input-output relationship of the model and not on the specific implementation of the model. IG integrates gradients along a path from a baseline to the actual input, solving the gradient saturation issue and adhering to these two axioms, thus laying a strong foundation for subsequent attribution methods~\cite{sundararajan2017axiomatic}.

SmoothGrad (SG) further extends IG by adding noise to the input and calculating gradients multiple times to reduce noise, generating clearer explanations~\cite{smilkov2017smoothgrad}. However, while SG improves interpretability, it increases computational costs and may miss subtle features critical to the decision process. Guided-IG~\cite{kapishnikov2021guided} and ExpectedGrad~\cite{erion2021improving} further reduce noise and improve the stability of explanations by guiding gradient integration paths and sampling multiple times. In addition, Boundary Integrated Gradients (BIG) focuses on gradient analysis in specific decision boundary regions, making it particularly effective in classification tasks~\cite{wang2021robust}. Fast-IG accelerates the attribution process by reducing the number of gradient computations, dynamically adjusting step sizes, and leveraging parallel computing~\cite{hesse2021fast}.

\subsection{Adversarial Example-Based Attribution Methods}
Adversarial example-based attribution methods build upon gradient-based methods by revealing model vulnerabilities through subtle input perturbations. These methods provide deeper explanations by analyzing how perturbations in input features impact model predictions. For instance, Adversarial Gradient Integrations (AGI) have been used to explain the effects of adversarial attacks on model decisions, helping to understand model robustness~\cite{pan2021explaining}. However, AGI may focus excessively on the model’s sensitivity to small perturbations, neglecting critical features of normal inputs.

To further improve explanation accuracy and efficiency, more faithful boundary-based attribution methods, such as MFABA, utilize higher-order derivatives and explore decision boundaries to provide more fine-grained and accurate explanations, though at a higher computational cost~\cite{zhu2023mfaba}. AttEXplore explores model parameters and decision boundary transitions to explain attribution but faces high computational demands, limiting its application on large-scale datasets~\cite{zhuattexplore}. Iterative Search Attribution (ISA) introduces scaling parameters during the iterative process to ensure that features gradually gain higher importance, enhancing the hierarchical nature of attributions~\cite{zhuiterative}. Local Attribution (LA) combines adversarial attacks with local spatial exploration, effectively optimizing the issue of ineffective intermediate states and providing clearer explanations for model predictions~\cite{zhu2024enhancing}.

\subsection{Adversarial Attacks} 
Adversarial attacks are based on fine adjustments to input data that cause the model to make incorrect predictions and can be classified into white-box and black-box attacks. White-box attacks assume that the attacker has access to internal model information such as gradients and weights, while black-box attacks assume that the attacker can only query the model's input-output relations. Gradient-based attacks are one of the classical methods, with FGSM being a representative technique that computes the gradient of the input with respect to the loss function and adjusts the input along the gradient direction to cause significant changes in model outputs~\cite{goodfellow2014explaining}. Although FGSM is computationally simple and easy to implement, the adversarial examples it generates are relatively coarse and easy to detect. PGD attacks extend FGSM by performing multi-step iterations to optimize adversarial perturbations, generating more concealed and effective adversarial examples that can bypass most defense mechanisms, though at the cost of higher computational complexity and longer attack times~\cite{madry2017towards}. Optimization-based attacks treat adversarial example generation as an optimization problem, aiming to minimize perturbations while maximizing the loss function. C\&W attacks represent this type of method, which generates adversarial examples by optimizing perturbation size and classification loss, making them harder to detect and capable of bypassing most defense mechanisms, albeit at a high computational cost and requiring adjustment for different models~\cite{carlini2017towards}.

Black-box attacks do not rely on internal model information. Transfer attacks are a common black-box attack method, based on the principle that adversarial examples generated on one model can also be effective on another, thus achieving cross-model attacks. Such attacks do not rely on the target model's internal information but instead infer potential adversarial perturbations through multiple queries of the model's input-output relationships, making them particularly useful when internal model information is unavailable. Many methods aim to improve the transferability of adversarial examples, as highly transferable adversarial examples can help attribution methods verify their interpretability and robustness across different models, revealing model vulnerabilities in similar features, thereby improving the reliability and generalizability of attribution analysis~\cite{jin2024benchmarking}. In this paper, we utilize the following transferability methods to optimize the performance of the AttEXplore+ framework: Momentum Iterative Method (MIM) enhances the transferability of adversarial examples across models by introducing a momentum mechanism~\cite{dong2018boosting}; Momentum Integrated Gradients (MIG) combines integrated gradient attribution and momentum strategies to further improve the cross-model generalizability of adversarial examples~\cite{ma2023transferable}; Diverse Input Method (DIM) performs multi-scale random scaling on input images, ensuring that adversarial examples exhibit good transferability across different network architectures~\cite{xie2019improving}; Scale-Invariant Nesterov Iterative Method (SI-NIM) introduces scale-invariance, allowing adversarial examples to maintain strong transferability across images of different sizes and resolutions~\cite{lin2019nesterov}; Translation-Invariant Method (TIM) adopts gradient-based convolution operations to optimize perturbation directions, significantly improving the cross-model transferability of adversarial examples~\cite{dong2019evading}; Neuron Attribution-based Attack (NAA) analyzes neuron attribution information to attack important neurons, achieving more efficient transfer attacks across models~\cite{zhang2022improving}; Structure Invariant Attack (SIA) generates adversarial examples with higher transferability while maintaining image structure through structure-invariant random image transformations~\cite{wang2023structure}; Gradient Relevance Attack (GRA) is a classical gradient-based method; however, its transferability heavily relies on hyperparameters for selecting neighborhood information~\cite{zhu2023boosting}; Frequency-based Stationary Point Search (FSPS) combines frequency domain analysis with stationary point search to significantly improve the cross-model transferability of adversarial examples by optimizing frequency information~\cite{zhu2023improving}. These methods optimize the transferability of adversarial examples from different angles, enhancing their cross-model attack capabilities.

\section{Preliminaries}

\subsection{Problem Definition}\label{PF}
Given a deep neural network model $f(\cdot)$ and an input sample $x \in R^n$, where $R^n$ denotes an $n$-dimensional sample space, the goal of attribution is to compute the contribution $A_i$ of each feature in the input sample $x$ to the model's decision $f(x;y)$. In the context of image classification, $y$ represents the predicted class of the model.

\subsection{Integrated Gradients (IG)}
In attribution methods, gradient information plays a crucial role in measuring feature importance. According to the concept of Saliency Maps~\cite{simonyan2013deep}, if $f(\cdot)$ is continuously differentiable, the gradients $\frac{\partial f}{\partial x}$ can be used to compute $A_i$, which is the importance of the $i$-th dimension in the input $x$. Integrated Gradients (IG)~\cite{sundararajan2017axiomatic} computes the attribution by integrating the gradients along a straight-line path from a baseline sample $x'$ to the input sample $x$. The gradients accumulated along this path are referred to as integrated gradients. For the $i$-th feature in the input sample, IG provides the contribution $A_i$ as shown in Equation~\ref{eqig}:

\begin{equation}
\label{eqig}
A_i:=(x_{i}-x_{i}')\int_{\alpha=0}^{1} \frac{\partial f(x'+\alpha \times(x-x'))}{\partial x_{i}}\mathrm{d}\alpha
\end{equation}

Here, $\frac{\partial f(\cdot)}{\partial x_{i}}$ represents the gradient of the deep neural network model $f(\cdot)$ along the $i$-th dimension. It is important to note that the baseline sample $x'$ should be selected based on the specific model being explained. For instance, for image models, $x'$ can be a black image, while for text models, it can be a zero embedding vector. The term $x'+\alpha \times(x-x')$ represents the linear path of IG. Since directly integrating the gradients from $x$ to $x'$ is difficult, IG approximates the integral by dividing the path into sufficiently small intervals and summing the gradients at each point along the path. The approximate computation of IG is given by Equation~\ref{eqigapp}:

\begin{equation}
\label{eqigapp}
A_i:=(x_{i}-x_{i}')\sum_{k=1}^{m}  \frac{\partial f(x'+\frac{k}{m} \times(x-x'))}{\partial x_{i}}\times \frac{1}{m}
\end{equation}

\subsection{The Relationship between Adversarial Attacks and Attribution} \label{advattack}

From the IG attribution process, we understand that IG requires a baseline sample as the starting point of integration. However, the choice of baseline sample is arbitrary and lacks strict justification. Although choosing a black image as the baseline works for simple tasks (such as MNIST classification), selecting a suitable baseline for more complex tasks significantly impacts the interpretability of the deep neural network model~\cite{pan2021explaining}. If we assume that the selected baseline sample always alters the model's output, the attribution problem becomes equivalent to finding the key features that change the model's decision. Since adversarial examples can be generated on different deep neural network models and easily mislead them, the IG integration process can be transformed into an integration process along the gradient path from adversarial examples to input samples.

It is noteworthy that adversarial examples can be defined as baseline samples given a specific deep neural network model, eliminating the ambiguity and inconsistency caused by selecting a baseline sample for a particular task. Next, we provide a detailed definition of adversarial attacks.

\textit{Definition of Adversarial Attack}: From Sec.~\ref{PF}, we know that given an input sample $x$ with a true label $y$, the prediction output of a deep neural network model with parameters $\theta$ is defined as $f(x;y;\theta)$. Suppose we have an adversarial sample $x_{adv}$, then the output becomes $f(x_{adv};\theta) = y'$, where $y \ne y'$. Clearly, the features that cause the shift from $y$ to $y'$ play a crucial role in the model's decision. The goal of adversarial attacks is to construct an adversarial sample $x_{adv}$ that belongs to the $\epsilon$-ball $\Phi_{\epsilon}(x)$ surrounding the input sample $x$, where $\Phi_{\epsilon}(x)=\left \{ x_{ball}:\left \| x_{ball}-x  \right \|_{p} \le \epsilon  \right  \}$. Here, $\epsilon$ is the perturbation upper bound that ensures the fidelity of the adversarial sample. $\left \| \cdot \right \|_{p}$ denotes the $L_p$ norm constraint, such as the $L_2$ norm. Let $L(x;y)$ be the loss function of the deep neural network model, which measures the difference between the true label $y$ and the predicted output $f(x;y;\theta)$. Adversarial attacks can be viewed as solving the following optimization problem:

\begin{equation}
    \max_{x_{adv}\in \Phi_{\epsilon}(x)} L(x_{adv};y) 
\end{equation}

As shown in Equation~\ref{bimit}, to mislead the model more stably, the Basic Iterative Method (BIM)~\cite{kurakin2018adversarial} iterates over the adversarial sample under a given perturbation limit:

\begin{equation}
\begin{aligned} 
  \label{bimit}
g(x^t) &= \bigtriangledown_{x_{adv}^{t}}L(x_{adv}^{t};y) \\
    x^{t+1}_{adv}&=\Pi_{\Phi_{\epsilon}(x)}\left \{ x_{adv}^{t}+\alpha \cdot sign(g(x^t))   \right \} 
\end{aligned}
\end{equation}

Here, $t$ represents the $t$-th iteration, and $\alpha$ is the learning rate. Assuming there are a total of $T$ iterations, then $\alpha=\frac{\epsilon}{T}$. $sign(\cdot)$ denotes the sign function, which determines the direction of gradient updates. Note that when $t=0$, $x^{0}_{adv}$ is the input sample $x$. The mapping function $\Pi_{\Phi_{\epsilon}(x)}$ is used to project the adversarial sample back into the $\epsilon$-ball $\Phi_{\epsilon}(x)$ to ensure that the perturbation does not exceed the allowable limit $\epsilon$. The step size $\alpha=\frac{\epsilon}{T}$ is chosen to ensure that the total perturbation after $T$ iterations remains within the perturbation limit $\epsilon$, thus maintaining the adversarial sample inside the $\epsilon$-ball. State-of-the-art attribution methods, such as Adversarial Gradient Integrations (AGI)~\cite{pan2021explaining} and parameter-exploring explanations like AttEXplore~\cite{zhuattexplore}, integrate the gradients corresponding to adversarial examples along the adversarial attack path after each iteration. Since the gradient update direction in each iteration is not necessarily aligned with the previous iteration, the adversarial attack path is often non-linear. This characteristic distinguishes it from the IG attribution path and provides a better explanation for the non-linear parts of deep neural networks. However, we find that in real-world scenarios, it is often difficult to access the parameters $\theta$ of deep neural network models. Currently, promising adversarial examples can be constructed in black-box settings using transfer-based attacks~\cite{dong2018boosting,dong2019evading,long2022frequency}, but the attribution performance of these adversarial examples has not been fully explored. In the next section, we will further analyze the profound impact of adversarial examples constructed by different transfer-based attacks on attribution.

% In this section, we first review the AttEXplore algorithm. Then, we derive the details of AttEXplore more rigorously and explore the effects of different transfer-based attacks on AttEXplore. Finally, we provide the pseudocode for AttEXplore++ in Algorithm~\ref{alg1}.
\section{Exploring the Impact of Adversarial Transferability on Attribution Performance}

This section explores the influence of adversarial transferability on attribution performance. It first reviews the AttEXplore method, focusing on how gradient information is obtained from adversarial attacks. Then, different adversarial attack methods are compared in terms of their gradient update strategies and their impact on transferability. Lastly, the role of randomness in adversarial attacks is analyzed, highlighting its effects on robustness, reproducibility, and attribution outcomes.

\subsection{AttEXplore Review}
The attribution result for adversarial sample $x^{t}$ is defined as:
\begin{equation}
\begin{aligned}
g(x^t) &= \bigtriangledown_{x^{t}}L(x^{t};y) \\
    A_i:&=\int\bigtriangleup x^{t} \cdot g(x^t) \mathrm{d}t 
\label{eq2}
\end{aligned}
\end{equation}

\begin{figure}[htpb]
    \centering
    \includegraphics[width=\linewidth]{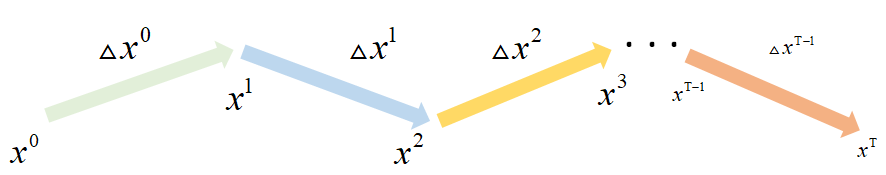}
\caption{Non-linear Attribution Path} 
\label{path} 
\end{figure}

In Equation~\ref{eq2}, $x^{t}$ is the adversarial baseline sample, while $x^{0}$ is the starting point of the attribution process, and $g(x^t)$ represents the gradient information, which can be obtained using different adversarial attack methods. This paper focuses on exploring the feasibility of using adversarial attacks with transferability to obtain gradient information, as discussed in Sec.~\ref{advattack}. The gradient iterates along a non-linear path $x^{t}=x^{0}+\sum_{k=0}^{t-1}\bigtriangleup x^{k}$, where $\triangle x^{t}$ represents the change in the sample along the adversarial attack path. For each iteration, the attribution method can employ the BIM method~\cite{kurakin2018adversarial} to update the adversarial sample as the baseline point. Thus, as shown in Figure~\ref{path}, we can derive $\bigtriangleup x^{t}=\alpha \cdot sign(\bigtriangledown_{x^{t}}L(x^{t};y))$.

\subsection{Different Ways to Obtain Gradient Information}
Transfer-based attacks do not directly train adversarial examples on the target deep neural network model but rather on a substitute model. This requires the adversarial examples trained on the substitute model to reliably cross the decision boundary of the target model and mislead its decision. Furthermore, in training adversarial examples, the proportion of training data in the total sample space is small, leading to issues with the decision boundary (as most of the data is in-distribution), and the boundaries are often rough (since the training samples are far from the decision boundary, adversarial training can make the decision boundary rough). Therefore, the minimal perturbations we obtain are not robust. Our goal is to train and cross more general decision boundaries, not just cross them.

\begin{table*}[]
\centering
\caption{Comparison of Gradient-Based Adversarial Attack Methods, Their Formulas, and Symbol Definitions.}
\label{tab.g}
\resizebox{.9\linewidth}{!}{%
\begin{tabular}{@{}c|c|c@{}}
\toprule
\textbf{Method} & \textbf{Formula for $g(x^{(t+1)})$}                                                                                         & \textbf{Meaning of Symbols} \\ \midrule
PGD~\cite{goodfellow2014explaining}             & $g(x^{(t+1)}) = \nabla_x L(f(x_{f}^{(t)}), y)$                                                                              & $\nabla_x L$: gradient of loss $L$ with respect to input $x$ \\
                &                                                                                                                              & $f$: model \\
                &                                                                                                                              & $x_f^{(t)}$: input at iteration $t$ \\
                &                                                                                                                              & $y$: true label \\ \midrule
MIM~\cite{dong2018boosting}             & $g(x^{(t+1)}) = \mu \cdot g(x^{(t)}) + \frac{\nabla_x L(f(x_{f}^{(t)}), y)}{|\nabla_x L(f(x_{f}^{(t)}), y)|_1}$             & $\mu$: momentum factor \\
                &                                                                                                                              & $g(x^{(t)})$: gradient at previous iteration $t$ \\ \midrule
MIG~\cite{dong2018boosting}             & $g(x^{(t+1)}) = \mu \cdot g(x^{(t)}) + \frac{IG(f,x_{f}^{(t)},y)}{|IG(f,x_{f}^{(t)},y)|_1}$                                 & $IG$: Integrated Gradients \\ \midrule
DIM~\cite{xie2019improving}             & $g(x^{(t+1)}) = \mathbb{E}_{\rho} [\nabla_x L(f(D_\rho(x_{f}^{(t)})), y)]$                                                  & $\mathbb{E}_{\rho}$: expectation over transformation $\rho$ \\
                &                                                                                                                              & $D_\rho$: input transformation \\ \midrule
SI-NIM~\cite{lin2019nesterov}          & $g(x^{(t+1)}) = \mu \cdot g_t + \frac{1}{m} \sum_{i=0}^{m-1} \frac{\nabla_x L(f(S_i(x_{nest})), y)}{\|\nabla_x L(f(S_i(x_{nest})), y)\|_1}$ & $S_i$: scaling operation \\
                &                                                                                                                              & $x_{nest}$: Nesterov forward jump point \\ \midrule
TIM~\cite{dong2019evading}             & $g(x^{(t+1)}) = \mathbb{E}_{\tau} [\nabla_x L(f(T_\tau(x_{f}^{(t)})), y)]$                                                  & $T_\tau$: translation operation \\ \midrule
NAA~\cite{zhang2022improving}             & $g(x^{(t+1)}) = \mu \cdot g(x^{(t)}) + \frac{\nabla_x W_{A_\phi}}{|\nabla_x W_{A_\phi}|_1}$                                 & $W_{A_\phi}$: weighted neuron attribution \\ \midrule
SIA~\cite{wang2023structure}             & $g(x^{(t+1)}) = \mathbb{E}_{T_n} [\nabla_x L(f(T_n(x_{f}^{(t)})), y)]$                                                      & $T_n$: transformation process \\ \midrule
GRA~\cite{zhu2023boosting}             & $g(x^{(t+1)}) = \mu \cdot g_t + \frac{1}{m} \sum_{i=1}^{m} \left( \frac{ G_t(x) \cdot G_i(x) }{ \| G_t(x) \|_2 \cdot \| G_i(x) \|_2 } \cdot G_t \right.$ & $G_t$: gradient at time $t$ \\

                & $\left. + \left( 1 - \frac{ G_t(x) \cdot G_i(x) }{ \| G_t(x) \|_2 \cdot \| G_i(x) \|_2 } \right) \cdot G_i \right)$  & $G_i$: gradient at different iteration \\ 
                &                                                                                                                              & $G_i(x)$: sampled via $\gamma_{it} \sim U\left[ -(\beta \cdot \epsilon)^d, (\beta \cdot \epsilon)^d \right]$ \\ 
                &                                                                                                                              & $\gamma_{it}$: The neighborhood perturbation \\
                &                                                                                                                              & $\beta$: The control factor of the perturbation magnitude, introducing randomness \\ \midrule
FSPS~\cite{zhu2023improving}            & $g(x^{(t+1)}) = \mu \cdot g_t + \frac{1}{n} \sum_{i=1}^{n} \frac{\nabla_x L(f(x_{\text{idct}}^{(i)}), y)}{\|\nabla_x L(f(x_{\text{idct}}^{(i)}), y)\|_1}$                                                          & $x_{\text{idct}}$: input transformed by inverse DCT \\
                &                                                                                                                              & $n$: number of frequency bands \\
\bottomrule
\end{tabular}% 
}
\end{table*}

In essence, the nature of transfer-based attacks is to explore the possibility of adversarial examples crossing decision boundaries. Although AttEXplore~\cite{zhuattexplore} explores more general decision boundaries by incorporating frequency domain information, the contribution of different adversarial example construction methods to attribution performance has not been fully explored. Different adversarial attack methods generate adversarial examples using various strategies and handle gradient updates $g(x^{(t+1)})$ differently. In Table~\ref{tab.g}, for the first time, we integrate transfer-based attack principles such as gradient editing, semantic similarity, and target modification into the attribution process to explore the impact of adversarial example transferability on attribution performance. Specifically, the PGD method directly updates perturbations through gradient descent and clips the result after each step. In contrast, MIM introduces a momentum term to smooth gradient updates and normalizes the gradient using the $L_1$ norm to prevent the direction from being dominated by the gradient magnitude. MIG combines integrated gradients with momentum, smoothing the gradients and enhancing stability through normalization.

DIM enhances perturbation diversity by applying random scaling and cropping operations without affecting semantics. Similarly, SI-NIM incorporates noise at each iteration, further enhancing the robustness of adversarial examples. TIM improves the attack efficacy of adversarial examples at different locations through translation transformations.

\subsection{The Impact of Randomness}

In the adversarial attack methods mentioned earlier, the setting of parameters plays a significant role in introducing randomness during the perturbation generation process. First, parameters related to diversity probabilit, $DP$ , introduce noise during gradient computation in methods such as DIM, TIM, and AttEXplore to enhance privacy protection. In each gradient update, the original gradient \( g(x) \) is perturbed by adding noise \( \eta \), with the update formula as:

\[
g'(x) = g(x) + \eta
\]

where \( \eta \sim \mathcal{N}(0, \sigma^2) \) is Gaussian noise. The introduction of this noise causes uncertainty in the perturbation direction, making the generated adversarial examples behave differently across different experimental runs. Therefore, differential privacy-related parameters not only affect the magnitude of perturbations but also introduce randomness to the method, impacting the stability and reproducibility of adversarial example generation.

Second, the parameter \( \beta \) in GRA is used to control the strength of regularization or the step size for gradient updates. The update formula for gradient descent is:

\[
x_{t+1} = x_t - \beta \nabla f(x_t)
\]

When randomness is introduced into the setting of \( \beta \), the magnitude and direction of perturbations will vary, leading to uncertainty in adversarial example generation under different experimental conditions. This may affect the model's robustness and the reproducibility of attribution results.

Moreover, the parameter \( \epsilon \) in methods such as PGD and SIA is a standard parameter used to control the perturbation upper bound, restricting the maximum difference between the adversarial and original samples. The perturbation constraint is given by:

\[
x' = x + \delta \quad \text{where} \quad \|\delta\|_\infty \leq \epsilon
\]

Although \( \epsilon \) itself is not a random parameter, the initial perturbation \( \delta_0 \) and step size may introduce randomness during each iteration, making the adversarial example generation process stochastic.

Lastly, the parameters \( \rho \) and \( \sigma \) used in FSPS and AttEXplore control the magnitude and variance of noise. Specifically, the added noise \( \eta \) follows:

\[
\eta \sim \mathcal{N}(0, \sigma^2)
\]

Adjusting \( \rho \) and \( \sigma \) can change the strength and range of noise, thereby affecting the randomness of the generated adversarial examples. Under different experimental conditions, the configuration of this noise will lead to significant differences in adversarial example behavior. Therefore, these noise control parameters largely determine the randomness and robustness of the attack methods.

The impact of this randomness is important in attribution analysis because the core goal of attribution analysis is to explain the decision-making process of the model. Randomness may lead to different perturbation paths for the same input in different experiments, affecting the stability of attribution results. Moreover, randomness may impact the reproducibility of experiments, making it difficult to replicate results, thereby affecting the credibility of scientific research. Finally, excessive randomness may make attribution results overly sensitive to input perturbations, weakening the robustness of the explanations. Therefore, studying the impact of randomness on attribution methods helps us identify and control uncertainty factors, improving the robustness and interpretability of attribution analysis.

% Please add the following required packages to your document preamble:
% \usepackage{booktabs}
% \usepackage{graphicx}
% \usepackage[normalem]{ulem}
% \useunder{\uline}{\ul}{}
\begin{table*}[]
\centering
    \caption{Comparison of insertion and deletion scores between AttEXplore++ and other interpretability methods across different models. The best insertion scores are in bold, and the second-best are underlined.}
    \label{tab:results}
\resizebox{\textwidth}{!}{%
\begin{tabular}{@{}c|cc|cc|cc|cc|cc@{}}
\toprule
Model        & \multicolumn{2}{c|}{Inception-v3} & \multicolumn{2}{c|}{ResNet-50} & \multicolumn{2}{c|}{VGG16} & \multicolumn{2}{c|}{MaxViT-T} & \multicolumn{2}{c}{ViT-B/16} \\ \midrule
Method       & Insertion           & Deletion    & Insertion         & Deletion   & Insertion       & Deletion & Insertion         & Deletion  & Insertion        & Deletion  \\ \midrule
BIG          & 0.3565              & 0.0360      & 0.2270            & 0.0402     & 0.1753          & 0.0289   & 0.5272            & 0.1762    & 0.4221           & 0.0928    \\
DeepLIFT     & 0.2182              & 0.0296      & 0.1033            & 0.0268     & 0.0701          & 0.0161   & 0.4856            & 0.1698    & 0.2964           & 0.0632    \\
EG           & 0.3035              & 0.3006      & 0.3237            & 0.3003     & 0.1951          & 0.1788   & 0.5239            & 0.4573    & 0.3607           & 0.3293    \\
FIG          & 0.1431              & 0.0322      & 0.0875            & 0.0296     & 0.0610          & 0.0197   & 0.4341            & 0.1713    & 0.2157           & 0.0709    \\
GIG          & 0.2266              & 0.0225      & 0.1251            & 0.0168     & 0.0787          & 0.0129   & 0.5423            & 0.1171    & 0.3354           & 0.0461    \\
IG           & 0.2274              & 0.0265      & 0.1121            & 0.0230     & 0.0688          & 0.0156   & 0.5301            & 0.1715    & 0.3458           & 0.0513    \\
MFABA        & 0.3961              & 0.0400      & 0.2576            & 0.0458     & 0.2145          & 0.0299   & 0.4197            & 0.3072    & 0.3651           & 0.1253    \\
SG           & 0.2933              & 0.0206      & 0.2330            & 0.0180     & 0.1407          & 0.0134   & \textbf{0.6432}   & 0.1119    & 0.4282           & 0.0346    \\
SM           & 0.1976              & 0.0296      & 0.1229            & 0.0315     & 0.0775          & 0.0207   & 0.4810            & 0.1759    & 0.3731           & 0.1246    \\
AGI          & 0.4230              & 0.0424      & 0.3792            & 0.0450     & 0.2725          & 0.0308   & 0.6079            & 0.1655    & 0.4251           & 0.0692    \\
AttEXplore   & {\ul 0.4614}        & 0.0288      & {\ul 0.4116}      & 0.0310     & {\ul 0.3032}    & 0.0206   & 0.5882            & 0.1293    & {\ul 0.4724}     & 0.0610    \\ \midrule
AttEXplore++ & \textbf{0.5335}     & 0.0496      & \textbf{0.4377}   & 0.0353     & \textbf{0.3494} & 0.0239   & {\ul 0.6368}      & 0.1750    & \textbf{0.4977}  & 0.0795    \\ \bottomrule
\end{tabular}%
}
\end{table*}

Furthermore, when investigating the impact of adversarial attacks on attribution analysis, it is important to explore how many iterations of white-box attacks succeed, as well as the few additional iterations performed after the attack succeeds. After a white-box attack succeeds, attackers may continue to perform additional attack operations to enhance the transferability of adversarial examples, ensuring that they retain high misclassification effectiveness even on black-box models.

Specifically, in the early stages of adversarial attacks (e.g., methods like PGD, MIM), the main goal of white-box attacks is to quickly find a perturbation \( \delta \) that effectively misleads the model such that \( f(x + \delta) \neq y \). Once the white-box model is successfully misled, the subsequent attack steps are not aimed at further improving the attack effectiveness on the white-box model but rather at increasing the generalization ability of the adversarial examples so that they can also successfully mislead the target model with different architectures or parameter settings. To achieve this, strategies such as random scaling (DIM), momentum updates (MIM), and noise injection (SI-NIM) are introduced. These strategies aim to enhance the transferability of adversarial examples by expanding the search space for perturbations and stabilizing gradient updates.

When studying these attack methods, analyzing the few attack steps executed after the white-box attack succeeds is crucial to understanding the mechanism by which adversarial example transferability is improved. For example, in DIM (multi-scale attacks), after the attack succeeds, random scaling of input images at different scales allows the attacker to obtain gradient information across multiple input scales, generating adversarial examples with better generalization ability. Similarly, MIM smooths gradient updates by introducing momentum terms, preventing transferability from being impaired due to excessive gradient fluctuations during white-box attacks.

Thus, these post-success attack steps are not merely optimizations of existing perturbations but rather strategies to ensure that the generated adversarial examples maintain high transferability across models. These extended attack strategies enhance the robustness of adversarial examples, giving them greater cross-model attack capabilities. The significance of these additional steps lies in their ability to allow attribution analysis results to extend beyond the decision-making process of a single model. They verify the robustness and generalizability of attribution methods by observing the cross-model performance of adversarial examples.

% Please add the following required packages to your document preamble:
% \usepackage{booktabs}
% \usepackage{graphicx}
% \usepackage[normalem]{ulem}
% \useunder{\uline}{\ul}{}
\begin{table*}[]
\centering
    \caption{Comparison of insertion and deletion scores between AttEXplore+ and different adversarial attack methods. The best insertion scores are in bold, and the second-best are underlined.}
    \label{tab:transfer}
\resizebox{\textwidth}{!}{%
\begin{tabular}{@{}c|cc|cc|cc|cc|cc@{}}
\toprule
Model            & \multicolumn{2}{c|}{Inception-v3} & \multicolumn{2}{c|}{ResNet-50} & \multicolumn{2}{c|}{VGG16} & \multicolumn{2}{c|}{MaxViT-T} & \multicolumn{2}{c}{ViT-B/16} \\ \midrule
Method           & Insertion           & Deletion    & Insertion         & Deletion   & Insertion       & Deletion & Insertion         & Deletion  & Insertion        & Deletion  \\ \midrule
AttEXplore+BIM   & 0.4056              & 0.0437      & 0.2713            & 0.0441     & 0.2162          & 0.0314   & 0.4567            & 0.2810    & 0.3555           & 0.1143    \\
AttEXplore+PGD   & 0.4069              & 0.0385      & 0.3015            & 0.0316     & 0.2422          & 0.0230   & {\ul 0.5403}      & 0.2178    & 0.3669           & 0.1013    \\
AttEXplore+MIM   & 0.4186              & 0.0604      & 0.2855            & 0.0567     & 0.2241          & 0.0375   & 0.4065            & 0.3520    & 0.3255           & 0.1272    \\
AttEXplore+DIM   & 0.4315              & 0.0612      & 0.3265            & 0.0647     & 0.2424          & 0.0444   & 0.3960            & 0.3744    & 0.3598           & 0.1779    \\
AttEXplore+SIA   & 0.3952              & 0.0585      & 0.2867            & 0.0485     & 0.2107          & 0.0318   & 0.4545            & 0.2901    & 0.3610           & 0.1451    \\
AttEXplore+SINIM & 0.4601              & 0.0383      & 0.3398            & 0.0404     & 0.2762          & 0.0286   & 0.4154            & 0.3312    & 0.3570           & 0.0817    \\
AttEXplore+TIM   & 0.4366              & 0.0524      & 0.3511            & 0.0571     & 0.2736          & 0.0397   & 0.4116            & 0.3808    & 0.3868           & 0.1627    \\
AttEXplore+MIG   & \textbf{0.5335}     & 0.0496      & {\ul 0.4348}      & 0.0366     & \textbf{0.3494} & 0.0239   & 0.5251            & 0.2996    & \textbf{0.4977}  & 0.0795    \\
AttEXplore+GRA   & {\ul 0.4968}        & 0.0419      & \textbf{0.4377}   & 0.0353     & {\ul 0.3257}    & 0.0244   & \textbf{0.6363}   & 0.1760    & {\ul 0.4096}     & 0.0555    \\ \bottomrule
\end{tabular}%
}
\end{table*}

\begin{figure*}[t]
    \centering
    \caption{Evaluation of the impact of different random seeds on the insertion and deletion scores of different adversarial attack methods in AttEXplore+. The results show that randomness has a limited impact on attribution performance, exhibiting high stability.}
    \label{fig:randomness}
    \includegraphics[width=1.0\textwidth]{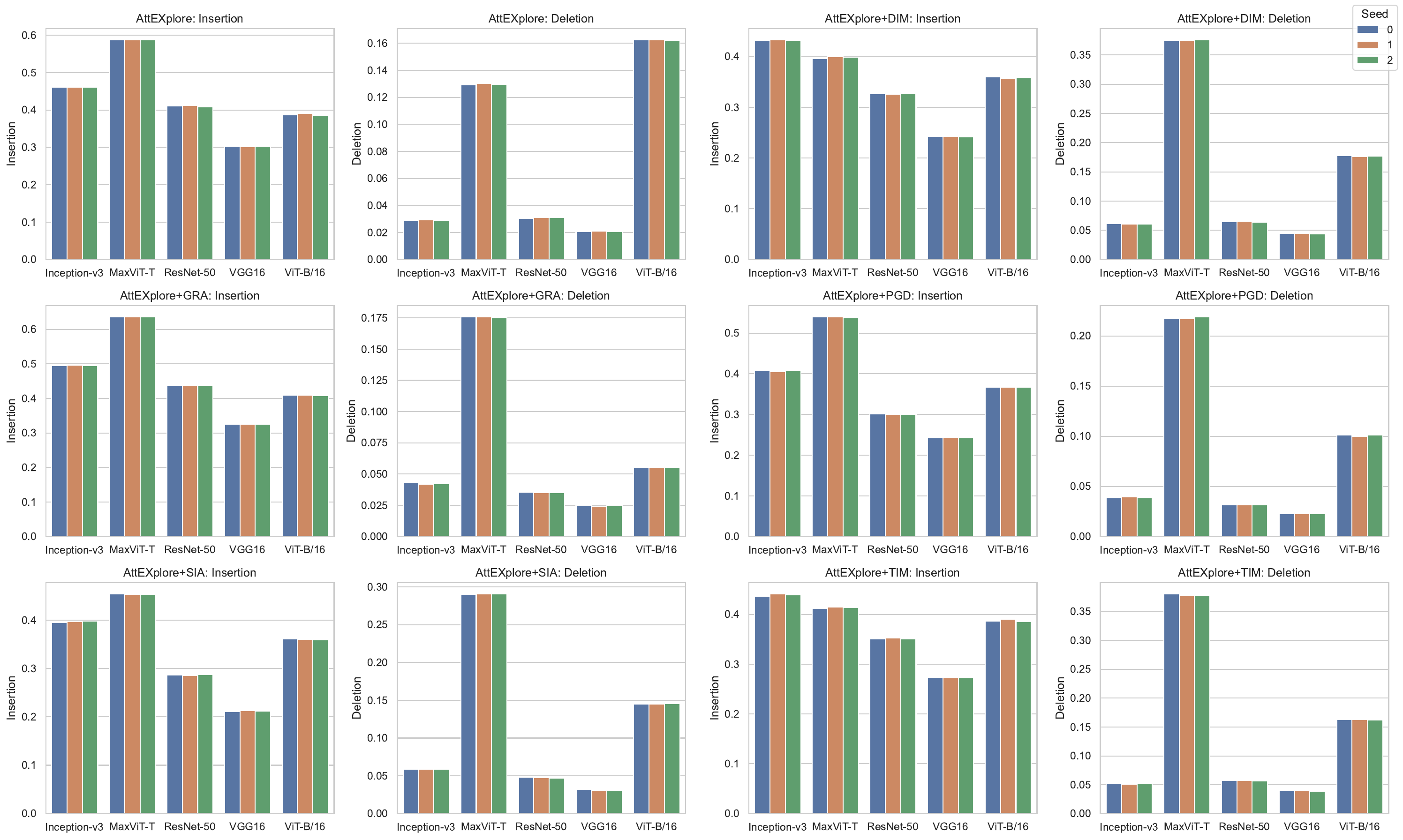}
\end{figure*}

\begin{figure*}[t]
    \centering
    \caption{The impact of diversity probability $DP$ on the insertion and deletion scores of AttEXplore and its variants across different models. Insertion and deletion scores are represented by blue and orange, respectively.}
    \label{fig:dp_impact}
    \includegraphics[width=1.0\textwidth]{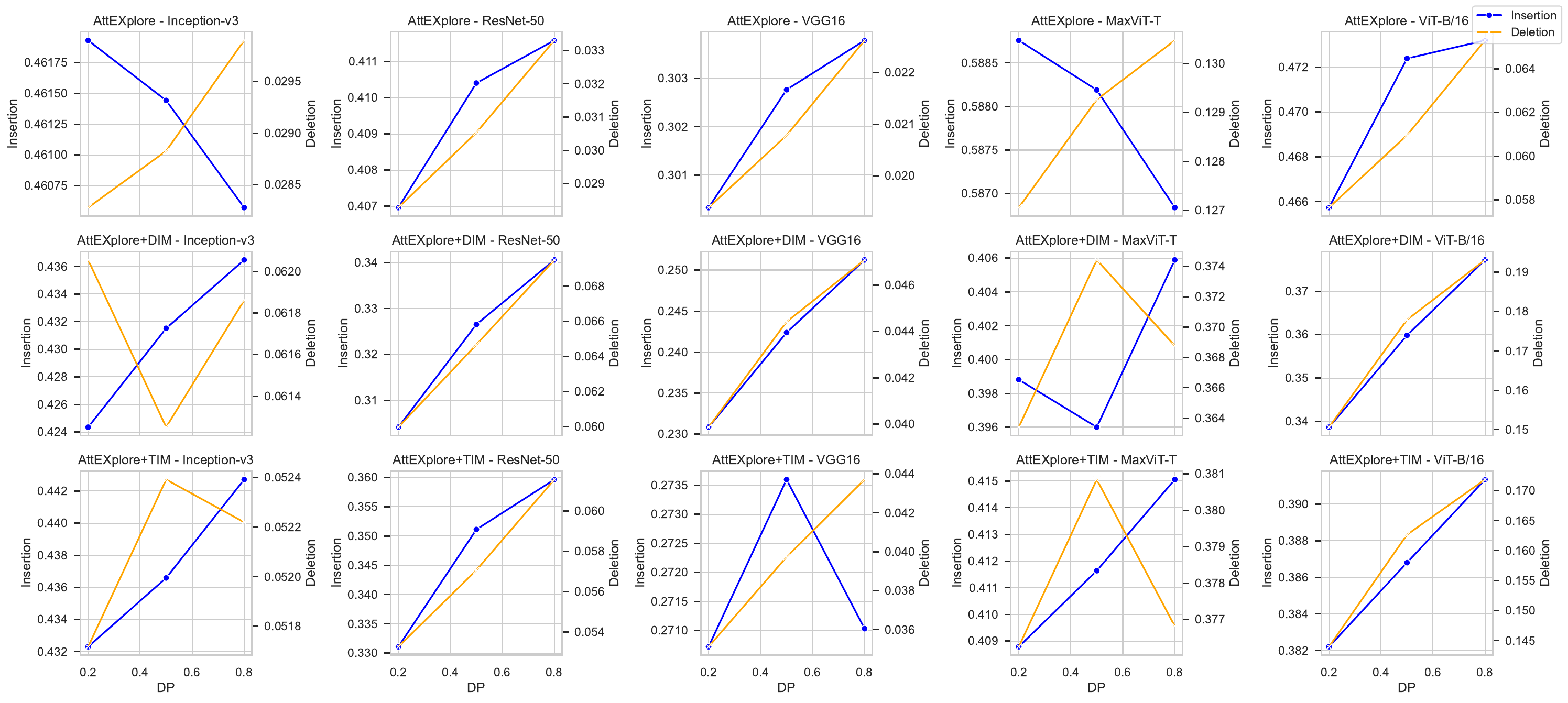}
\end{figure*}

\section{Experiments}
\subsection{Datasets and Models}
In this study, we conduct experiments on the ImageNet dataset~\cite{deng2009imagenet}. Specifically, we extract 1000 samples from the ImageNet dataset and evaluate them using various existing interpretability methods and adversarial attack methods~\cite{zhang2022improving,long2022frequency,pan2021explaining,zhu2023mfaba,zhuattexplore,zhuiterative}. We select three classical Convolutional Neural Networks (CNN) models: Inception-v3~\cite{szegedy2016rethinking}, ResNet-50~\cite{he2016deep}, and VGG16~\cite{simonyan2014very}. Additionally, we include the MaxViT-T and ViT-B/16~\cite{dosovitskiy2020image} models to explore the interpretability of the proposed methods on vision transformer-based models.

\subsection{Baseline Methods}
We select 11 classical and state-of-the-art interpretability algorithms as baseline methods to compare against our proposed AttEXplore++ method. These algorithms include BIG, DeepLIFT~\cite{shrikumar2017learning}, EG~\cite{erion2021improving}, FIG~\cite{hesse2021fast}, GIG~\cite{kapishnikov2021guided}, IG~\cite{sundararajan2017axiomatic}, MFABA, SG~\cite{smilkov2017smoothgrad}, SM~\cite{simonyan2013deep}, AGI~\cite{pan2021explaining}, and AttEXplore. Among them, AttEXplore serves as the primary comparison for our proposed method.

\subsection{Evaluation Metrics}
We use insertion score and deletion score, commonly adopted for evaluating interpretability algorithms, as the evaluation metrics~\cite{pan2021explaining}. The insertion score measures the change in the model output when important pixels are gradually added back to the input. Higher scores indicate better interpretability of the algorithm. The deletion score assesses the change in the model output when pixels are gradually removed, where lower scores suggest stronger interpretability. These two metrics reduce reliance on manual labeling and perform well in evaluating causal explanations without human bias. In attribution methods, the insertion score is typically more important than the deletion score, as the adversarial characteristics of neural networks may lead to misleading results in deletion scores~\cite{petsiuk2018rise}. Therefore, the insertion score is regarded as the primary metric for attribution performance, while the deletion score serves as an auxiliary dimension for analysis.

\subsection{Experimental Settings}
All experiments were conducted on a Linux platform using two Nvidia A100 GPUs. In the AttEXplore+ framework, we chose the most optimal strategy to build our AttEXplore++ method. Specifically, in the Inception-v3 and ViT-B/16 models, we utilized MIG to optimize the transferability of adversarial attacks to improve the attribution process. For ResNet-50, VGG16, and MaxViT-T models, we applied the GRA method to enhance the attribution process. While generating adversarial examples, we set the noise amplitude $\beta$ to 4.0. For the AttEXplore framework, we set the suppression factor $\rho$ to 0.5, perturbation rate $\epsilon$ to 16, and diversity probability $DP$ to 0.5.

\subsection{Experimental Results}
As shown in Table~\ref{tab:results}, we comprehensively compare AttEXplore++ with other interpretability methods on five models using insertion scores and deletion scores. AttEXplore++ demonstrates superior performance across all models. Compared to the second-best AttEXplore, AttEXplore++ improves overall performance by an average of 7.57\%; specifically, it improves by an average of 8.17\% on classical CNN models (Inception-v3, ResNet-50, and VGG16) and 6.36\% on transformer-based models (MaxViT-T and ViT-B/16). Compared to all competing algorithms, AttEXplore++ shows an average performance improvement of 32.62\%; more precisely, it improves by 38.02\% on CNN models and 21.82\% on transformer models. These results indicate that utilizing more transferable adversarial attack methods in the attribution process can significantly enhance the performance of attribution algorithms. We will further discuss this point in the next section.

% \begin{table}[h]
%     \centering
%     \caption{Comparison of insertion and deletion scores between AttEXplore++ and other interpretability methods across different models. The best insertion scores are in bold, and the second-best are underlined.}
%     \label{tab:results}
%     \includegraphics[width=1.0\textwidth]{images/image.png}
% \end{table}

\subsection{Impact of Transferability on Interpretability}
Table~\ref{tab:transfer} presents the impact of different transferable adversarial attack methods on the attribution performance within the AttEXplore+ framework. It is evident that with the enhancement of attack method transferability, the attribution performance of AttEXplore+ significantly improves. Specifically, traditional white-box attack methods like BIM and PGD exhibit weaker interpretability, while more recent adversarial attack methods like MIG and GRA demonstrate the best performance, achieving the best or second-best results across nearly all models. For instance, the MIG method achieves the best insertion scores of 0.5335 and 0.4348 on the Inception-v3 and ResNet-50 models, respectively, while the GRA method achieves the best insertion score of 0.4977 on the ViT-B/16 model. These results indicate that stronger transferability plays a key role in enhancing the performance of attribution algorithms.

% \begin{table}[h]
%     \centering
%     \caption{Comparison of insertion and deletion scores between AttEXplore+ and different adversarial attack methods. The best insertion scores are in bold, and the second-best are underlined.}
%     \label{tab:transfer}
%     \includegraphics[width=1.0\textwidth]{images/image.png}
% \end{table}

\subsection{Impact of Randomness on Interpretability}

As shown in Figure~\ref{fig:randomness}, we evaluate the impact of randomness on different attribution methods. In the experiments, we fix the diversity probability ($DP$) to 0.5, the noise amplitude $\beta$ to 4.0, and the perturbation rate $\epsilon$ to 16. We tested the impact of randomness on attribution performance using three different random seeds. The results show that randomness has minimal impact on the insertion and deletion scores of most methods. In almost all methods, the variation in attribution performance due to changes in random seeds is very small, with the insertion and deletion scores remaining stable. This demonstrates that the AttEXplore+ framework is robust to randomness, ensuring the stability and reliability of the attribution performance.

\begin{figure*}[t]
    \centering
    \caption{The impact of perturbation rate $\epsilon$ on the insertion and deletion scores of AttEXplore and its variants across different models. Insertion and deletion scores are represented by blue and orange, respectively.}
    \label{fig:epsilon_impact}
    \includegraphics[width=1.0\textwidth]{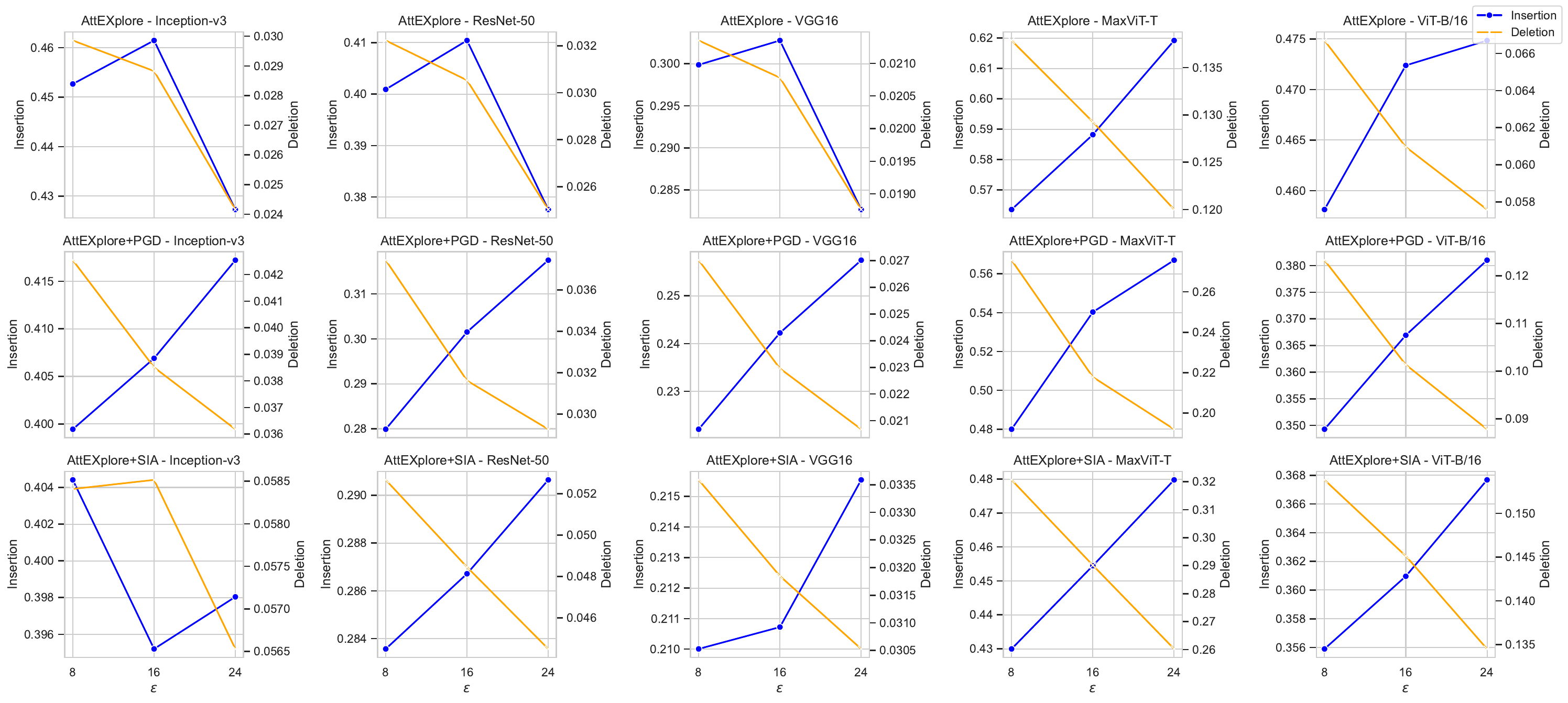}
\end{figure*}

\begin{figure*}[t]
    \centering
    \caption{The impact of noise amplitude $\beta$ on the insertion and deletion scores of the AttEXplore+GRA method across different models. Insertion and deletion scores are represented by blue and orange, respectively.}
    \label{fig:beta_impact}
    \includegraphics[width=1.0\textwidth]{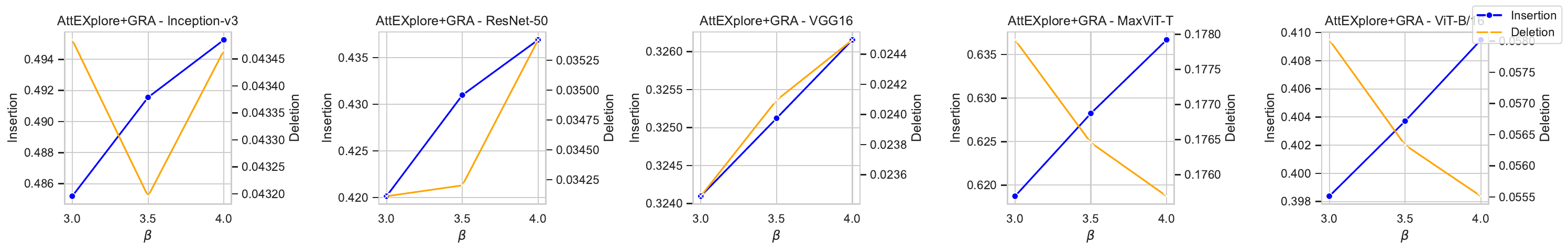}
\end{figure*}

\subsection{Impact of Transferability Method Parameters on Attribution Performance}

This section examines the impact of the diversity probability ($DP$) on the attribution performance of AttEXplore and its two variants (AttEXplore+DIM and AttEXplore+TIM). We fix $\rho$ at 0.5, the perturbation rate $\epsilon$ at 16.0, and the random seed at 0, and test $DP$ values of 0.2, 0.5, and 0.8. As shown in Figure~\ref{fig:dp_impact}, different $DP$ values have varying impacts on different models. On ResNet-50, VGG16, and ViT-B/16, both the insertion and deletion scores increase synchronously with increasing $DP$. Given that insertion scores are significantly higher than deletion scores in magnitude and more critical for attribution evaluation, higher $DP$ values are recommended for these models.

In contrast, on Inception-v3 and MaxViT-T models, the insertion scores of the original AttEXplore decrease as $DP$ increases, while the deletion scores increase, indicating that larger $DP$ values lead to performance degradation in these models. However, when AttEXplore's attribution process is optimized using DIM and TIM, the insertion scores generally increase with increasing $DP$, while the deletion scores show an initial increase followed by a decrease. Overall, after optimization with DIM and TIM, the attribution performance of AttEXplore improves on these models.

\subsection{Impact of Perturbation Rate $\epsilon$ on AttEXplore+ Performance}

This section examines the impact of the perturbation rate $\epsilon$ on the performance of three methods (AttEXplore, AttEXplore+PGD, and AttEXplore+SIA). In the experiments, we fix $\rho$ at 0.5, the random seed at 0, and set $DP$ to 0.5, testing $\epsilon$ values of 8, 16, and 24. As shown in Figure~\ref{fig:epsilon_impact}, in most cases, the insertion scores increase with increasing $\epsilon$, while the deletion scores decrease, indicating that interpretability improves with higher perturbation rates.

It is worth noting that the interpretability of the original AttEXplore decreases on traditional CNN models as $\epsilon$ increases. However, after optimizing AttEXplore with adversarial attack methods, the insertion scores significantly improve, and the deletion scores decrease with increasing $\epsilon$, indicating a marked improvement in interpretability.

\subsection{Impact of Noise Amplitude $\beta$ on AttEXplore+ Performance}

In this section, we evaluate the impact of noise amplitude $\beta$ on the performance of AttEXplore+ when optimized with GRA. We conduct experiments with $\beta$ values of $3.0$, $3.5$, and $4.0$. As shown in Figure~\ref{fig:beta_impact}, the insertion scores are positively correlated with noise amplitude $\beta$, as insertion scores increase with $\beta$ across all models. For deletion scores, the scores also increase with increasing $\beta$ in traditional CNN models. However, in transformer-based models, the interpretability improves optimally as $\beta$ increases, as shown by an increase in insertion scores and a gradual decrease in deletion scores. This indicates that larger noise amplitudes help improve attribution performance in these models.

\section{Conclusion}
In this work, we proposed the AttEXplore+ framework, which integrates multiple adversarial attack methods to explore the relationship between adversarial attack transferability and attribution interpretability. By selecting the most effective adversarial strategies, we introduced AttEXplore++, an optimized method that significantly improves interpretability compared to prior approaches. Our experiments demonstrate that AttEXplore++ enhances the performance of attribution methods across various models, including CNNs and vision transformers, with substantial improvements in insertion and deletion scores. Furthermore, we examined key parameters such as perturbation rate, noise amplitude, and diversity probability, showing that AttEXplore++ is robust against randomness, delivering stable and reliable explanations. These results underscore the critical role of adversarial transferability in improving attribution methods, offering deeper insights into neural network decision-making processes and paving the way for further advancements in model interpretability.

\bibliography{main}

\begin{thebibliography}{10}

\bibitem{lecun2015deep}
Y.~LeCun, Y.~Bengio, and G.~Hinton, ``Deep learning,'' {\em nature}, vol.~521, no.~7553, pp.~436--444, 2015.

\bibitem{esteva2017dermatologist}
A.~Esteva, B.~Kuprel, R.~A. Novoa, J.~Ko, S.~M. Swetter, H.~M. Blau, and S.~Thrun, ``Dermatologist-level classification of skin cancer with deep neural networks,'' {\em nature}, vol.~542, no.~7639, pp.~115--118, 2017.

\bibitem{chen2016xgboost}
T.~Chen and C.~Guestrin, ``Xgboost: A scalable tree boosting system,'' in {\em Proceedings of the 22nd acm sigkdd international conference on knowledge discovery and data mining}, pp.~785--794, 2016.

\bibitem{doshi2017towards}
F.~Doshi-Velez and B.~Kim, ``Towards a rigorous science of interpretable machine learning,'' {\em arXiv preprint arXiv:1702.08608}, 2017.

\bibitem{lipton2018mythos}
Z.~C. Lipton, ``The mythos of model interpretability: In machine learning, the concept of interpretability is both important and slippery.,'' {\em Queue}, vol.~16, no.~3, pp.~31--57, 2018.

\bibitem{rudin2019stop}
C.~Rudin, ``Stop explaining black box machine learning models for high stakes decisions and use interpretable models instead,'' {\em Nature machine intelligence}, vol.~1, no.~5, pp.~206--215, 2019.

\bibitem{goodfellow2014explaining}
I.~J. Goodfellow, J.~Shlens, and C.~Szegedy, ``Explaining and harnessing adversarial examples,'' {\em arXiv preprint arXiv:1412.6572}, 2014.

\bibitem{mehrabi2021survey}
N.~Mehrabi, F.~Morstatter, N.~Saxena, K.~Lerman, and A.~Galstyan, ``A survey on bias and fairness in machine learning,'' {\em ACM computing surveys (CSUR)}, vol.~54, no.~6, pp.~1--35, 2021.

\bibitem{somani2024propagating}
A.~Somani, L.~A. Horsch, A.~Bopardikar, and D.~K. Prasad, ``Propagating transparency: A deep dive into the interpretability of neural networks,'' 2024.

\bibitem{hooker2019benchmark}
S.~Hooker, D.~Erhan, P.-J. Kindermans, and B.~Kim, ``A benchmark for interpretability methods in deep neural networks,'' {\em Advances in neural information processing systems}, vol.~32, 2019.

\bibitem{montavon2018methods}
G.~Montavon, W.~Samek, and K.-R. M{\"u}ller, ``Methods for interpreting and understanding deep neural networks,'' {\em Digital signal processing}, vol.~73, pp.~1--15, 2018.

\bibitem{tjoa2020survey}
E.~Tjoa and C.~Guan, ``A survey on explainable artificial intelligence (xai): Toward medical xai,'' {\em IEEE transactions on neural networks and learning systems}, vol.~32, no.~11, pp.~4793--4813, 2020.

\bibitem{thibeau2023interpretability}
E.~Thibeau-Sutre, S.~Collin, N.~Burgos, and O.~Colliot, ``Interpretability of machine learning methods applied to neuroimaging,'' {\em Machine Learning for Brain Disorders}, pp.~655--704, 2023.

\bibitem{narodytska2017simple}
N.~Narodytska and S.~P. Kasiviswanathan, ``Simple black-box adversarial attacks on deep neural networks.,'' in {\em CVPR Workshops}, vol.~2, 2017.

\bibitem{wang2022di}
Y.~Wang, J.~Liu, X.~Chang, R.~J. Rodr{\'\i}guez, and J.~Wang, ``Di-aa: An interpretable white-box attack for fooling deep neural networks,'' {\em Information Sciences}, vol.~610, pp.~14--32, 2022.

\bibitem{ross2018improving}
A.~Ross and F.~Doshi-Velez, ``Improving the adversarial robustness and interpretability of deep neural networks by regularizing their input gradients,'' in {\em Proceedings of the AAAI conference on artificial intelligence}, vol.~32, 2018.

\bibitem{subramanya2019fooling}
A.~Subramanya, V.~Pillai, and H.~Pirsiavash, ``Fooling network interpretation in image classification,'' in {\em Proceedings of the IEEE/CVF international conference on computer vision}, pp.~2020--2029, 2019.

\bibitem{zhang2021adversarial}
X.~Zhang, X.~Zheng, and W.~Mao, ``Adversarial perturbation defense on deep neural networks,'' {\em ACM Computing Surveys (CSUR)}, vol.~54, no.~8, pp.~1--36, 2021.

\bibitem{han2023interpreting}
S.~Han, C.~Lin, C.~Shen, Q.~Wang, and X.~Guan, ``Interpreting adversarial examples in deep learning: A review,'' {\em ACM Computing Surveys}, vol.~55, no.~14s, pp.~1--38, 2023.

\bibitem{zhuattexplore}
Z.~Zhu, H.~Chen, J.~Zhang, X.~Wang, Z.~Jin, J.~Xue, and F.~D. Salim, ``Attexplore: Attribution for explanation with model parameters exploration,'' in {\em The Twelfth International Conference on Learning Representations}, 2024.

\bibitem{ribeiro2016should}
M.~T. Ribeiro, S.~Singh, and C.~Guestrin, ``" why should i trust you?" explaining the predictions of any classifier,'' in {\em Proceedings of the 22nd ACM SIGKDD international conference on knowledge discovery and data mining}, pp.~1135--1144, 2016.

\bibitem{lundberg2017unified}
S.~M. Lundberg and S.-I. Lee, ``A unified approach to interpreting model predictions,'' {\em Advances in neural information processing systems}, vol.~30, 2017.

\bibitem{shrikumar2017learning}
A.~Shrikumar, P.~Greenside, and A.~Kundaje, ``Learning important features through propagating activation differences,'' in {\em International conference on machine learning}, pp.~3145--3153, PMLR, 2017.

\bibitem{sundararajan2017axiomatic}
M.~Sundararajan, A.~Taly, and Q.~Yan, ``Axiomatic attribution for deep networks,'' in {\em International conference on machine learning}, pp.~3319--3328, PMLR, 2017.

\bibitem{simonyan2013deep}
K.~Simonyan, A.~Vedaldi, and A.~Zisserman, ``Deep inside convolutional networks: Visualising image classification models and saliency maps,'' {\em arXiv preprint arXiv:1312.6034}, 2013.

\bibitem{smilkov2017smoothgrad}
D.~Smilkov, N.~Thorat, B.~Kim, F.~Vi{\'e}gas, and M.~Wattenberg, ``Smoothgrad: removing noise by adding noise,'' {\em arXiv preprint arXiv:1706.03825}, 2017.

\bibitem{kapishnikov2021guided}
A.~Kapishnikov, S.~Venugopalan, B.~Avci, B.~Wedin, M.~Terry, and T.~Bolukbasi, ``Guided integrated gradients: An adaptive path method for removing noise,'' in {\em Proceedings of the IEEE/CVF conference on computer vision and pattern recognition}, pp.~5050--5058, 2021.

\bibitem{erion2021improving}
G.~Erion, J.~D. Janizek, P.~Sturmfels, S.~M. Lundberg, and S.-I. Lee, ``Improving performance of deep learning models with axiomatic attribution priors and expected gradients,'' {\em Nature machine intelligence}, vol.~3, no.~7, pp.~620--631, 2021.

\bibitem{wang2021robust}
Z.~Wang, M.~Fredrikson, and A.~Datta, ``Robust models are more interpretable because attributions look normal,'' {\em arXiv preprint arXiv:2103.11257}, 2021.

\bibitem{hesse2021fast}
R.~Hesse, S.~Schaub-Meyer, and S.~Roth, ``Fast axiomatic attribution for neural networks,'' {\em Advances in Neural Information Processing Systems}, vol.~34, pp.~19513--19524, 2021.

\bibitem{pan2021explaining}
D.~Pan, X.~Li, and D.~Zhu, ``Explaining deep neural network models with adversarial gradient integration,'' in {\em Thirtieth International Joint Conference on Artificial Intelligence (IJCAI)}, 2021.

\bibitem{zhu2023mfaba}
Z.~Zhu, H.~Chen, J.~Zhang, X.~Wang, Z.~Jin, M.~Xue, D.~Zhu, and K.-K.~R. Choo, ``Mfaba: A more faithful and accelerated boundary-based attribution method for deep neural networks,'' {\em arXiv preprint arXiv:2312.13630}, 2023.

\bibitem{zhuiterative}
Z.~Zhu, H.~Chen, X.~Wang, J.~Zhang, Z.~Jin, J.~Xue, and J.~Shen, ``Iterative search attribution for deep neural networks,'' in {\em Forty-first International Conference on Machine Learning}, 2024.

\bibitem{zhu2024enhancing}
Z.~Zhu, Z.~Jin, J.~Zhang, and H.~Chen, ``Enhancing model interpretability with local attribution over global exploration,'' {\em arXiv preprint arXiv:2408.07736}, 2024.

\bibitem{madry2017towards}
A.~Madry, A.~Makelov, L.~Schmidt, D.~Tsipras, and A.~Vladu, ``Towards deep learning models resistant to adversarial attacks,'' {\em arXiv preprint arXiv:1706.06083}, 2017.

\bibitem{carlini2017towards}
N.~Carlini and D.~Wagner, ``Towards evaluating the robustness of neural networks,'' in {\em 2017 ieee symposium on security and privacy (sp)}, pp.~39--57, Ieee, 2017.

\bibitem{jin2024benchmarking}
Z.~Jin, J.~Zhang, Z.~Zhu, and H.~Chen, ``Benchmarking transferable adversarial attacks.,'' {\em CoRR}, 2024.

\bibitem{dong2018boosting}
Y.~Dong, F.~Liao, T.~Pang, H.~Su, J.~Zhu, X.~Hu, and J.~Li, ``Boosting adversarial attacks with momentum,'' in {\em Proceedings of the IEEE conference on computer vision and pattern recognition}, pp.~9185--9193, 2018.

\bibitem{ma2023transferable}
W.~Ma, Y.~Li, X.~Jia, and W.~Xu, ``Transferable adversarial attack for both vision transformers and convolutional networks via momentum integrated gradients,'' in {\em Proceedings of the IEEE/CVF International Conference on Computer Vision}, pp.~4630--4639, 2023.

\bibitem{xie2019improving}
C.~Xie, Z.~Zhang, Y.~Zhou, S.~Bai, J.~Wang, Z.~Ren, and A.~L. Yuille, ``Improving transferability of adversarial examples with input diversity,'' in {\em Proceedings of the IEEE/CVF conference on computer vision and pattern recognition}, pp.~2730--2739, 2019.

\bibitem{lin2019nesterov}
J.~Lin, C.~Song, K.~He, L.~Wang, and J.~E. Hopcroft, ``Nesterov accelerated gradient and scale invariance for adversarial attacks,'' {\em arXiv preprint arXiv:1908.06281}, 2019.

\bibitem{dong2019evading}
Y.~Dong, T.~Pang, H.~Su, and J.~Zhu, ``Evading defenses to transferable adversarial examples by translation-invariant attacks,'' in {\em Proceedings of the IEEE/CVF Conference on Computer Vision and Pattern Recognition}, pp.~4312--4321, 2019.

\bibitem{zhang2022improving}
J.~Zhang, W.~Wu, J.-t. Huang, Y.~Huang, W.~Wang, Y.~Su, and M.~R. Lyu, ``Improving adversarial transferability via neuron attribution-based attacks,'' in {\em Proceedings of the IEEE/CVF Conference on Computer Vision and Pattern Recognition}, pp.~14993--15002, 2022.

\bibitem{wang2023structure}
X.~Wang, Z.~Zhang, and J.~Zhang, ``Structure invariant transformation for better adversarial transferability,'' in {\em Proceedings of the IEEE/CVF International Conference on Computer Vision}, pp.~4607--4619, 2023.

\bibitem{zhu2023boosting}
H.~Zhu, Y.~Ren, X.~Sui, L.~Yang, and W.~Jiang, ``Boosting adversarial transferability via gradient relevance attack,'' in {\em Proceedings of the IEEE/CVF International Conference on Computer Vision}, pp.~4741--4750, 2023.

\bibitem{zhu2023improving}
Z.~Zhu, H.~Chen, J.~Zhang, X.~Wang, Z.~Jin, Q.~Lu, J.~Shen, and K.-K.~R. Choo, ``Improving adversarial transferability via frequency-based stationary point search,'' in {\em Proceedings of the 32nd ACM International Conference on Information and Knowledge Management}, pp.~3626--3635, 2023.

\bibitem{kurakin2018adversarial}
A.~Kurakin, I.~J. Goodfellow, and S.~Bengio, ``Adversarial examples in the physical world,'' in {\em Artificial intelligence safety and security}, pp.~99--112, Chapman and Hall/CRC, 2018.

\bibitem{long2022frequency}
Y.~Long, Q.~Zhang, B.~Zeng, L.~Gao, X.~Liu, J.~Zhang, and J.~Song, ``Frequency domain model augmentation for adversarial attack,'' in {\em European Conference on Computer Vision}, pp.~549--566, Springer, 2022.

\bibitem{deng2009imagenet}
J.~Deng, W.~Dong, R.~Socher, L.-J. Li, K.~Li, and L.~Fei-Fei, ``Imagenet: A large-scale hierarchical image database,'' in {\em 2009 IEEE conference on computer vision and pattern recognition}, pp.~248--255, Ieee, 2009.

\bibitem{szegedy2016rethinking}
C.~Szegedy, V.~Vanhoucke, S.~Ioffe, J.~Shlens, and Z.~Wojna, ``Rethinking the inception architecture for computer vision,'' in {\em Proceedings of the IEEE conference on computer vision and pattern recognition}, pp.~2818--2826, 2016.

\bibitem{he2016deep}
K.~He, X.~Zhang, S.~Ren, and J.~Sun, ``Deep residual learning for image recognition,'' in {\em Proceedings of the IEEE conference on computer vision and pattern recognition}, pp.~770--778, 2016.

\bibitem{simonyan2014very}
K.~Simonyan and A.~Zisserman, ``Very deep convolutional networks for large-scale image recognition,'' {\em arXiv preprint arXiv:1409.1556}, 2014.

\bibitem{dosovitskiy2020image}
A.~Dosovitskiy, ``An image is worth 16x16 words: Transformers for image recognition at scale,'' {\em arXiv preprint arXiv:2010.11929}, 2020.

\bibitem{petsiuk2018rise}
V.~Petsiuk, ``Rise: Randomized input sampling for explanation of black-box models,'' {\em arXiv preprint arXiv:1806.07421}, 2018.

\end{thebibliography}
\bibliographystyle{ieeetr}

\end{document}